\PassOptionsToPackage{usenames,dvipsnames,table}{xcolor}
\pdfoutput=1

\documentclass[11pt]{article}

\usepackage[]{EMNLP2023}

\usepackage{times}
\usepackage{latexsym}
\usepackage{booktabs}
\usepackage{caption}
\usepackage{subcaption}
\usepackage{listings}

\usepackage{pifont}
\newcommand{\cmark}{\textcolor{green!80!black}{\ding{51}}}
\newcommand{\xmark}{\ding{55}}
\usepackage{xcolor}
\usepackage{enumitem}

\usepackage[T1]{fontenc}

\usepackage[utf8]{inputenc}

\usepackage{microtype}

\usepackage{url}
\usepackage{amsmath}
\usepackage{amsfonts}
\usepackage{amssymb}
\usepackage{graphicx}
\usepackage{float}
\usepackage{lipsum}
\usepackage{multicol}
\usepackage{xcolor}
\usepackage{natbib}
\usepackage[font=small]{caption}
\usepackage{etoolbox,xspace}

\definecolor{backcolour}{rgb}{0.95,0.95,0.92}

\lstdefinestyle{codestyle}{
    backgroundcolor=\color{backcolour},   
    breakatwhitespace=false,         
    breaklines=true,         
    basicstyle=\ttfamily\scriptsize,
}
\newcommand{\dataset}{eyeStyliency\xspace}

\newcolumntype{P}[1]{>{\centering\arraybackslash}p{#1}}

\newtoggle{draft}
\togglefalse{draft}
\iftoggle{draft}{
\newcommand{\karin}[1]{\textcolor{magenta}{[#1 \textsc{--karin}]}}
\newcommand{\dk}[1]{\textcolor{cyan}{[#1 \textsc{--dk}]}}
\newcommand{\change}[1]{\textcolor{red}}
}{
\newcommand{\karin}[1]{}
\newcommand{\dk}[1]{}
}

\newcommand{\com}[1]{}

\title{
A Comparative Study on Textual Saliency of Styles \\from Eye Tracking, Annotations, and Language Models
}

\author{Karin de Langis \\
  University of Minnesota \\
  \texttt{dento019@umn.edu} \\\And
  Dongyeop Kang \\
  University of Minnesota \\
  \texttt{dongyeop@umn.edu} \\}

\begin{document}
\maketitle
	\begin{abstract}
There is growing interest in incorporating eye-tracking data and other implicit measures of human language processing into natural language processing (NLP) pipelines.
The data from human language processing contain unique insight into human linguistic understanding that could be exploited by language models. 
However, many unanswered questions remain about the nature of this data and how it can best be utilized in downstream NLP tasks. 
In this paper, we present \dataset, an eye-tracking dataset for human processing of stylistic text (e.g., politeness).
We develop a variety of methods to derive style saliency scores over text using the collected eye dataset. 
We further investigate how this saliency data compares to both human annotation methods and model-based interpretability metrics. 
We find that while eye-tracking data is unique, it also intersects with both human annotations and model-based importance scores, providing a possible bridge between human- and machine-based perspectives. We propose utilizing this type of data to evaluate the cognitive plausibility of models that interpret style.
Our eye-tracking data and processing code are publicly available.\footnote{\href{https://github.com/minnesotanlp/eyeStyliency}{https://github.com/minnesotanlp/eyeStyliency}}
\end{abstract}

\section{Introduction}
Human perception and understanding of text is critical in NLP. Typically, this understanding is leveraged in the form of ground-truth human annotations in supervised learning pipelines, or in the form of human evaluations of generated text. However, human language understanding is complex; multiple cognitive processes work together to enable reading, many of which occur automatically and unconsciously~\citep{devito1970psychology}. 

Because of the complexity, disciplines concerned with understanding and modeling how humans read -- e.g., psycholinguistics and cognitive science -- heavily utilize \textit{implicit} measures of the human reading experience that capture signals from these automatic processes in real time. Examples of implicit measures include event-related potential, reaction times, and eye movements. In contrast, \textit{explicit} measures include surveys and other methods that directly ask people to report their perceptions and experiences.  We posit that traditional NLP pipelines, which have widely used explicit measures of human understanding, can also benefit from implicit measures. In this paper, we focus specifically on \textit{the use of eye movements as an implicit measure} of textual saliency.

\begin{figure}[t]
    \centering
	\includegraphics[width=\columnwidth]{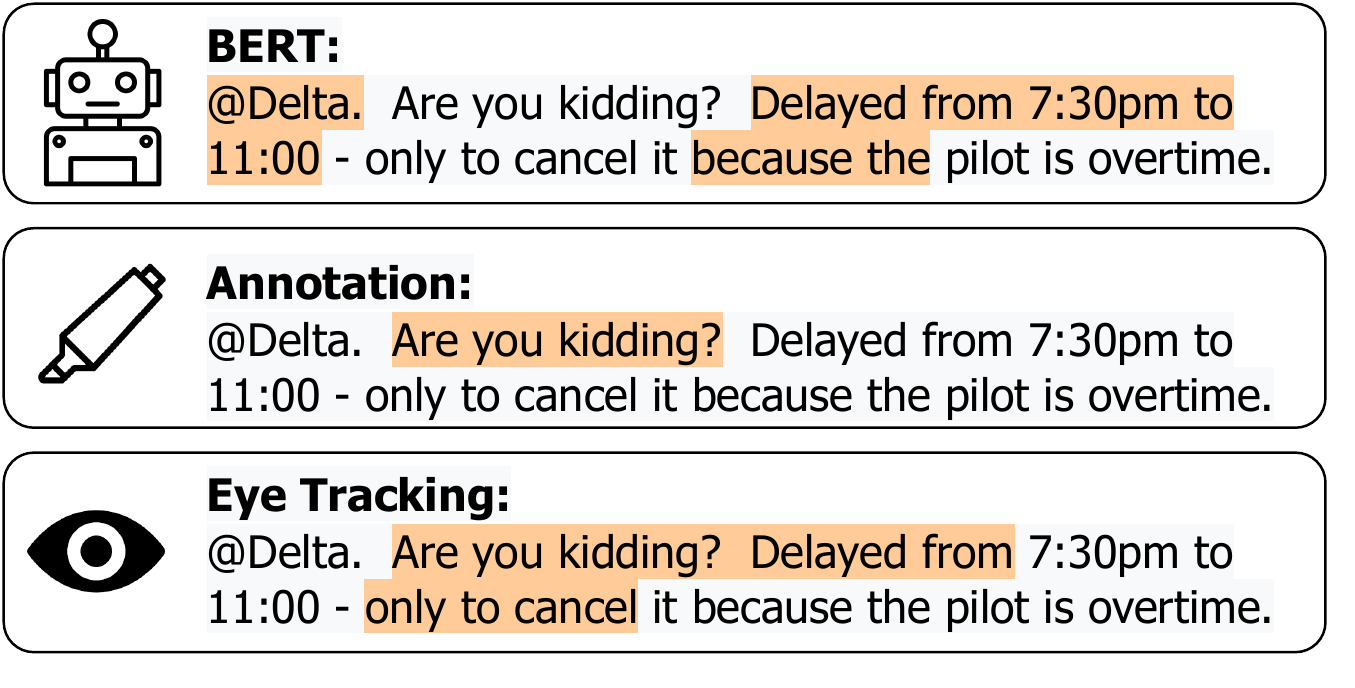}
	\caption{Salient words for \textit{impoliteness} from three different perspectives. We find that eye tracking data contains some overlap between machine and human-annotated salience.}
	\label{fig:fig1salience}
\end{figure}

Recent research in NLP has demonstrated the feasibility of incorporating various types of eye movement data into NLP models in order to improve performance on a number of tasks (see Table~\ref{tab:measures} for an overview). However, this is still an underexplored area: best practices remain unclear, and it's not obvious whether there are tasks that are unsuitable for eye movement data, or how eye movement data should be balanced with traditional annotation data. In this work, we address two main research questions: 
RQ1: Does eye-tracking-based saliency meaningfully differ from simply gathering word-level human annotations, or from model-based word importance measures?  RQ2: How can we measure eye movements specific to a high-level textual feature like style, and which eye tracking metrics and data processing methods are best suited to capturing textual saliency?

To address these questions, we conduct an eye tracking case study in which participants read texts the HummingBird dataset~\citep{hayati2021does}. We choose this dataset because it contains lexical-level human annotations indicating which words contribute to the text's style and because its domain (textual styles) has not to our knowledge been widely explored for eye tracking applications -- although prior work investigates eye tracking and sentiment analysis, it does not extend to other linguistic styles such as politeness.

We collect style-specific eye movements through a carefully designed experiment (see Section~\ref{methods} for details), and we use these eye movements to derive saliency scores over the text. We compare this eye-based saliency to human annotations as well as two large language model (LLM)-derived importance scores: 
integrated gradient scores from a BERT model fine-tuned on style datasets~\cite{hayati2021does}, and word-surprisal scores from GPT-2 \cite{radford2019language} (see Figure \ref{fig:fig1salience} for an example). Our findings indicate that eye-tracking-based saliency highlights some unique areas of the text, but it also intersects with both saliency from model-based metrics and saliency from human annotations, making a bridge of sorts between the human- and machine-based perspectives. We discuss some implications of these findings for NLP research. 

Specifically, our contributions are:
\begin{itemize}[noitemsep,topsep=0pt,leftmargin=7mm]
    \item An experimental paradigm for obtaining eye tracking-based signals for specific features of text (in our case, textual style).
    \item A first-of-its-kind eye movement dataset on style saliency, collected from 20 participants and consisting of both control readings and style-focused readings for polite, impolite, positive, and negative textual styles.
    \item An illustration of the distinction between this dataset's \textbf{explicit} human annotations and \textbf{implicit} human eye data through a unique comparison between salient text obtained via annotation and  via eye tracking.
\end{itemize}

\begin{table}[t]
\centering
\small
\begin{tabular}{@{}p{.18\textwidth}@{}P{.135\textwidth}@{}P{.025\textwidth}@{}P{.025\textwidth}@{}P{.11\textwidth}@{}}
\midrule
 & \textbf{NLP Area} & \textbf{H} & \textbf{M} & \textbf{learning from} \textbf{eye data} \\ \midrule

\textbf{Ours} & \begin{tabular}[c]{@{}c@{}}Textual \\Style\end{tabular} & \cmark & \cmark & \begin{tabular}[c]{@{}c@{}}\xmark \end{tabular} \\ \midrule
\begin{tabular}[c]{@{}c@{}}\citet{kuribayashi-etal-2021-lower}\end{tabular} & Perplexity & \xmark & \cmark & \xmark \\ \midrule
\citet{malmaud2020bridging} & QA & \xmark & \xmark & Joint learning \\ \midrule
\citet{bolotova2020people} & QA & \xmark & \cmark & \xmark \\ \midrule
\citet{sood2020improving} & QA & \xmark & \cmark & \xmark \\ \midrule
\citet{sood2020interpreting} & Paraphrasing & \xmark & \xmark & Joint learning \\ \midrule
\begin{tabular}[c]{@{}c@{}}\citet{hollenstein2019advancing}\end{tabular} & \begin{tabular}[c]{@{}c@{}}Sentiment \\Clf., NER \end{tabular}& \xmark & \xmark & \begin{tabular}[c]{@{}c@{}}Joint learning\end{tabular} \\ \midrule
\citet{barrett-etal-2018} & PoS tagging & \xmark & \xmark & HMM \\ \midrule
\citet{tokunaga2017eye} & NER & \xmark & \xmark & \xmark \\ \midrule
\citet{klerke2015looking} & Summarization & \cmark & \xmark & \xmark \\ \midrule
\citet{green2014eye}  & Parsing & \xmark & \xmark & \xmark \\ \midrule 
\end{tabular}
\caption{A summary of prior work applying eye tracking methods to NLP. The \textbf{H} column indicates whether traditional human annotations are considered in relation to the eye tracking data, and the \textbf{M} indicates whether model attention is considered. Most prior research has focused on either (a) comparing and contrasting eye movements with various models' attention mechanisms, or (b) using eye movements for multi-task learning, where NLP task performance can be improved by a model that jointly learns to predict eye movements in addition to the relevant NLP task. To our knowledge, there have not been three-way comparisons between attention mechanisms from eye tracking, large language models, and manual human annotations.}
\label{tab:cfs}
\end{table}

\section{Related Work}
Eye tracking has been a staple of psycholinguistic investigations of reading for decades~\citep{rayner1978eye, just1980theory}. Eye movement data is compelling because it provides realtime information about how people process language in a natural, ecologically valid setting (i.e., there is no explicit experimental task, such as question answering, for participants to complete) \citep{kaiser2013experimental}. Eye data provides insight into cognitive processes through the eye-mind assumption, which posits that (1) our eyes fixate on whatever our brains are currently processing, and (2) as cognitive effort to process an item increases, the amount of time that the eyes fixate on that item also increases \citep{just1980theory}. Analysis of eye data under this framework has led to important insights into many unconscious phenomena in human language comprehension, e.g. the mechanisms involved in ambiguity resolution during reading \citep{traxler2008role}.

\textbf{Eye Tracking in NLP}. Due to the eye-mind assumption, eye-tracking data is particularly well-suited to inferring patterns of reader attention, or saliency, over text. This saliency information has so far shown promising results when integrated into NLP models for question answering (e.g. \citet{malkin2021boosting, sood2020interpreting, malmaud2020bridging}). However, this is still a developing research area: there is limited available data, and there is little consensus regarding how to effectively collect data and incorporate it into NLP pipelines. To our knowledge there is no previous research that investigates saliency for style via eye tracking, nor any previous research that compares saliency from eye tracking to human annotations (Table \ref{tab:cfs} compares our work with the prior work). 

Outside of textual saliency, eye-tracking data has been leveraged for a variety of NLP tasks.  \citet{mishra2013automatically} quantify the difficulty of sentences in machine translation tasks using eye movement data;~\citet{mishra2016harnessing} determine whether a reader understands sarcasm in text, and~\citet{sogaard2016evaluating} evaluate the quality of word embeddings and text generations, respectively. Other work uses existing datasets, sometimes augmenting the data with a learned gaze predictor model, and uses this eye movement data as an additional signal when training models for various NLP tasks, including named entity recognition~\cite{hollenstein2019advancing,tokunaga2017eye}, paraphrasing~\cite{sood2020improving}, part-of-speech tagging~\cite{barrett-etal-2018}, and sentiment analysis (see also~\citet{mathias2020} for a review).

\textbf{Saliency in Linguistic Styles}. People apply styles to language in order to express attitudes, reflect interpersonal intentions or goals, or convey social standings of the speaker or listener. (Note that while many sociolinguistics theories distinguish between textual style and textual attributes, in this work, we follow the common convention in recent NLP papers of broadly using `style' to encompass both of these ideas \citep{jin-etal-2022-deep}.)  The meaning expressed by textual styles can be significant; in fact, there is strong evidence that effective communication requires an understanding of both style and literal semantic meaning~\cite{hovy1987generating}. 
Although BERT \cite{devlin2018bert} based fine-tuned models show strong performance on style classification, there are notable differences between how BERT perceives style at the lexical level and how humans perceive it, and that using data about these differences during training improves model performance \citep{hayati2023stylex}.

\section{\dataset: A Dataset of Eye Movement for Textual Saliency}\label{methods}
We describe the data collection procedure for \dataset dataset from 20 participants and methods for computing saliency scores over text. 

\subsection{Data Setups}
Our dataset consists of items from the Hummingbird dataset \cite{hayati2021does} in the following stylistic categories: \textit{polite}, \textit{impolite}, \textit{positive} \textit{sentiment}, and \textit{negative} \textit{sentiment}.\footnote{Politeness and sentiment datasets in Hummingbird are originally sourced from \citet{danescu2013computational} and \citet{socher2013recursive}.} We chose this subset because of the small correlation between categories (other categories, e.g. \textit{anger}, \textit{disgust}, and \textit{negative sentiment} are all highly correlated).  

In this study, we limit participants' total time commitment to one hour. To achieve this, the dataset size is 90 items across the four style categories. (The average word count per item in the dataset is 21.6 overall; for the impolite, polite, negative, and positive styles average word count is 21.3, 22.8, 21.4, and 20.8, respectively.) Most participants finished the experiment in 40-60 minutes, depending on both the individual's reading speed and the time needed to calibrate the individual to the eye tracker.


%
\begin{table*}[h]
\centering
\small
\begin{tabular}{@{}l@{}P{5.3cm}|P{0.5cm}|c|c|c|c|c|c|c}
\midrule
\multicolumn{1}{c}{} & \multicolumn{1}{c}{\textbf{Applications}}&\textbf{N} & \multicolumn{1}{c|}{\textbf{FFD}} & \multicolumn{1}{c|}{\textbf{FC}} & \multicolumn{1}{c|}{\textbf{FRD}} & \multicolumn{1}{c|}{\textbf{DT}} & \multicolumn{1}{c|}{\textbf{RR}} & \multicolumn{1}{c|}{\textbf{RC}} & \multicolumn{1}{c}{\textbf{PL}} \\ \midrule
\textbf{\dataset} (Ours) & Textual Style & 20  & \cmark & \xmark & \cmark & \cmark & \cmark & \xmark & \cmark \\ \midrule
\citet{kuribayashi-etal-2021-lower} & Language model perplexity & \xmark & \xmark & \xmark & \xmark & \cmark & \xmark & \xmark & \xmark \\\midrule
\citet{malmaud2020bridging} & Question Answering & 269 & \xmark & \xmark & \xmark & \cmark & \xmark & \xmark & \xmark   \\\midrule
\citet{bolotova2020people} & Question Answering & 20 & \xmark & \cmark & \xmark & \cmark & \cmark & \xmark & \xmark  \\\midrule
\citet{sood2020improving} & Paraphrasing & \xmark & \xmark & \cmark & \xmark & \xmark & \xmark & \xmark & \xmark  \\\midrule
\citet{sood2020interpreting} & Question Answering & 23 & \xmark & \xmark & \xmark & \cmark & \xmark & \xmark & \xmark  \\ \midrule
\citet{hollenstein2019advancing}& NER, Sentiment/Relation Classification & \xmark & \cmark & \cmark & \cmark & \cmark & \cmark & \cmark & \cmark  \\ \midrule
\citet{barrett-etal-2018} & PoS tagging & \xmark & \cmark & \xmark & \xmark & \cmark & \cmark & \cmark & \xmark \\ \midrule
\citet{tokunaga2017eye} & Named entity recognition (NER)  & \xmark & \xmark & \xmark & \xmark & \cmark & \xmark & \xmark & \xmark \\ \midrule
\citet{mishra2016harnessing} & Sarcasm detection & 7 & \xmark & \xmark & \xmark & \cmark & \xmark & \xmark & \xmark  \\ \midrule
\citet{klerke2015looking}  & NLG evaluation  & 24 & \xmark & \cmark & \xmark & \cmark & \cmark & \cmark & \cmark \\ \midrule
\citet{green2014eye} & Phrase-structure parsing & 40 & \xmark & \xmark & \xmark & \xmark & \cmark & \xmark & \xmark  \\ \midrule
\end{tabular}
\caption{A comparison of prior works with respect to the eye tracking metrics studied, data processing techniques, and number of participants whose eye tracking data is collected. FFD = first fixation duration, FC = fixation count, RC = regression count, RR = reread time, PL = pupil size, N = number of participants if new eye data collected.
}
\label{tab:measures}
\vspace{-3mm}
\end{table*}

\subsection{Eye-Tracking Measures}\label{subsec:measures}
Monocular eye movement data is collected with an EyeLink 1000 Plus\footnote{Made by SR Research, Ontario, Canada; \url{https://www.sr-research.com/eyelink-1000-plus/}} at a rate of 1000Hz. We look at the following eye-tracking metrics:

\begin{itemize}[noitemsep,topsep=0pt,leftmargin=7mm]
    \item First Fixation Duration (FFD): The duration of the first fixation in an interest area.
    \item First Run Dwell Time (FRD): The time interval beginning with the first fixation in the interest area and ending when the eye exits an interest area (whether to the right or left).
    \item Go Past Time (GP): The time interval beginning with the first fixation in an interest area and ending when the eye exits the interest area to the \textit{left} (i.e., to reread).
    \item Dwell Time (DT): The total fixation duration for all fixations in an interest area. Also known as gaze duration.
    \item Reread Time (RR): The total fixation duration for all fixations in an interest area after the area has already been entered and exited once.
    \item Pupil Size (PS): The average pupil size over all fixations in an interest area.
    
    (Note that First Run Dwell Time + Reread Time = Dwell Time.)
\end{itemize}

These measures can broadly be categorized into early measures (first fixation duration, pupil size) that reflect more low-level reading processes and late measures (go past time, dwell time, reread time) that reflect higher-level processing and meaning integration \cite{conklin-etal-2021-meta}. Previous eye tracking applications for NLP have commonly used dwell time, but a variety of measures have been examined (see Table~\ref{tab:measures}). In this study, we aim to compare a wide variety of measures in order to estimate which may be best-suited to capturing textual saliency. Note that to avoid redundancy, we chose to omit fixation counts from our analysis after finding high correlations between this measure and dwell time (pearson's $r = 0.93$, $p < 0.01$). We also chose to omit regression counts from our analysis after finding that regression counts were extremely sparse -- specifically, $1.8\%$ of the dataset had a non-zero regression count.

\subsection{Experimental Procedure}
\begin{figure}[t]
    \centering
	\includegraphics[width=\columnwidth,trim={0.1cm 0 0.1cm 0.1cm},clip]{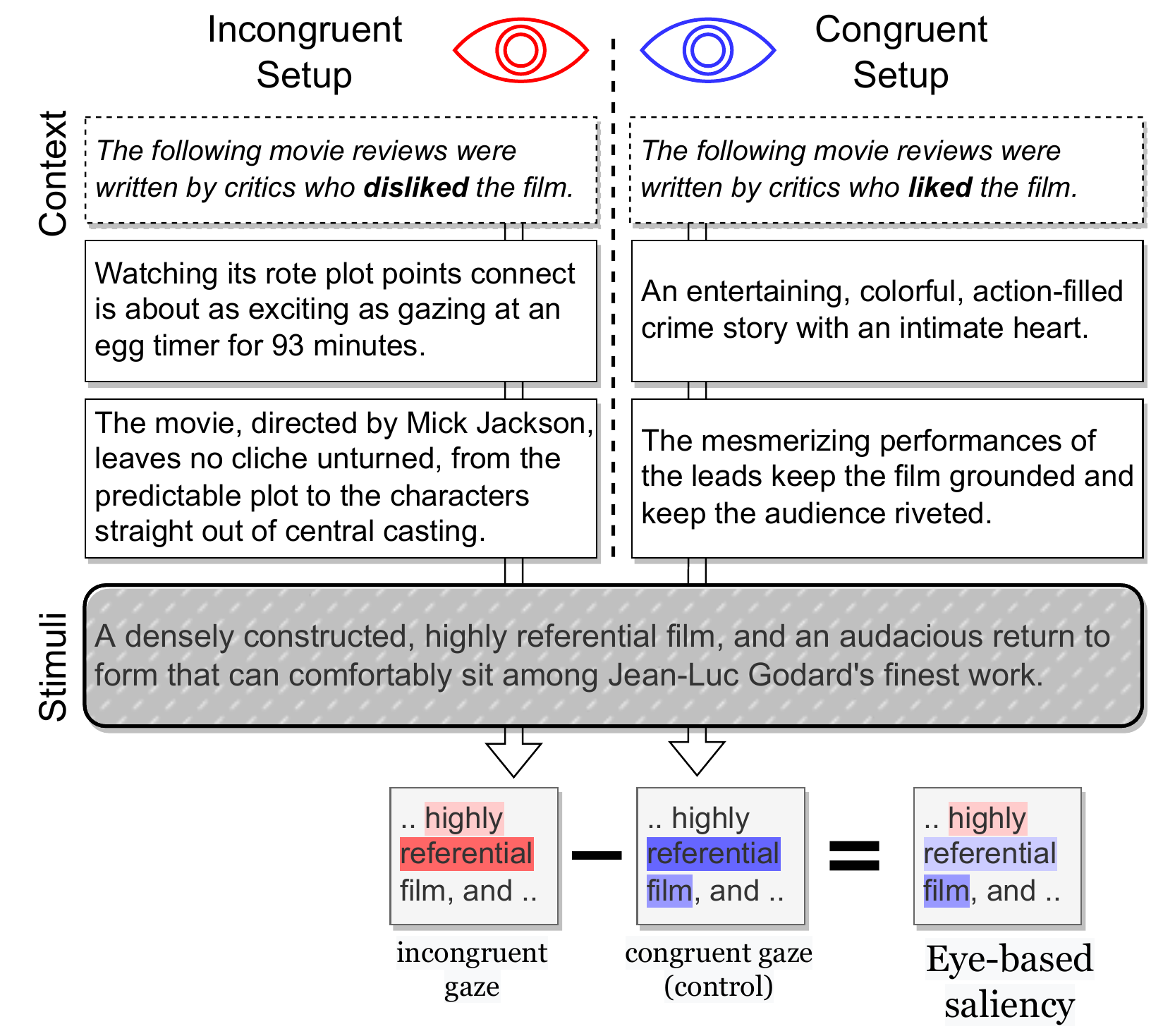}
	\caption{Illustrative example of congruent vs incongruent presentation of the same stimulus. We rely on expectation effects to induce participants to attend to the unexpected style (in this case, positive sentiment); in other words, we assume that the surprise regarding the style will result in longer gaze durations for words that contribute to the perception of that style --- in this case, words relating to positive sentiment.
    }
	\label{fig:con}
\end{figure}

\begin{figure}[t]
    \centering
	\includegraphics[width=\columnwidth]{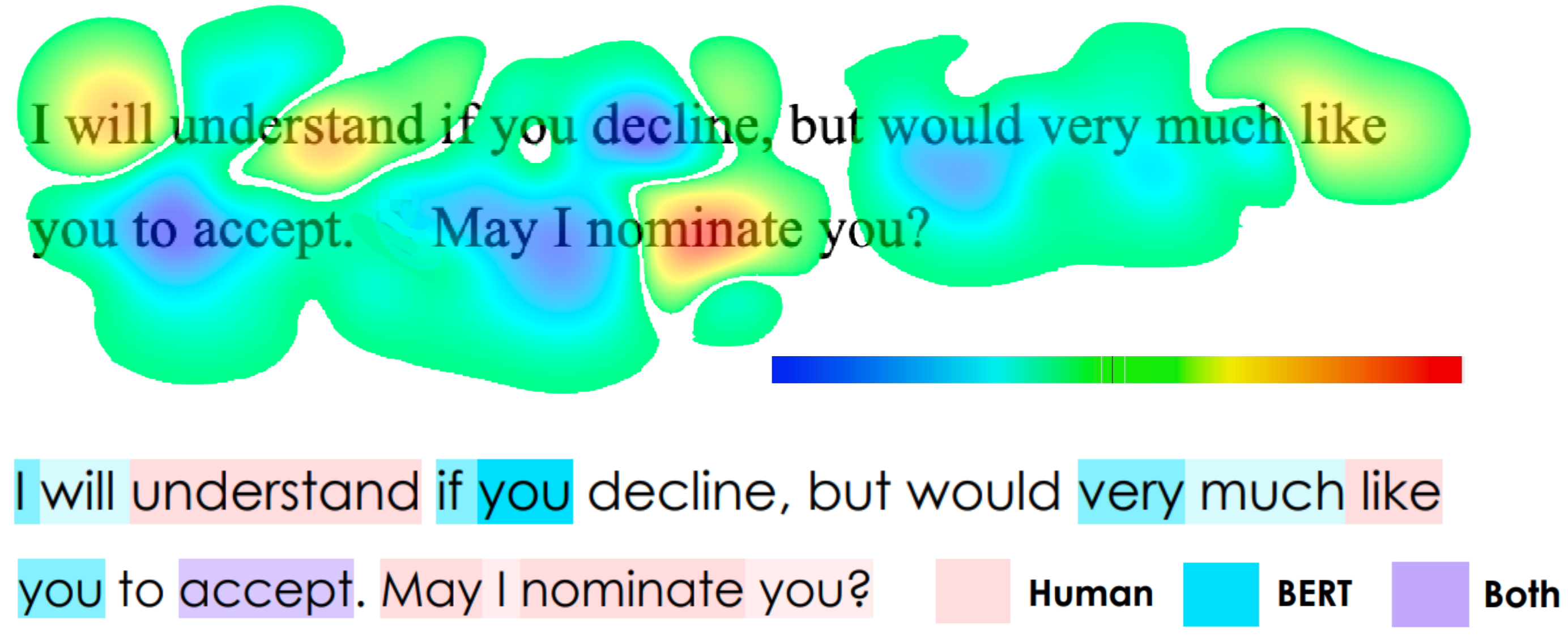}
	\caption{Exemplary eye-tracking data showing saliency for \textit{polite} style, with comparison to human word-level style importance highlighting. The eye-tracking data is visualized as a heat map showing gaze data from the incongruent style condition, with the gaze data from the congruent style (control) condition subtracted.}
	\label{fig:heatmap}
\end{figure}

The experiment follows a between-subjects, blocked design. The key part of our experiment is the technique to isolate eye movements that are specifically relevant to the text's style. In order to do this, we inform participants at the beginning of each block that the block will contain only stimuli that share a style (polite, impolite, positive, or negative) and source (Twitter, IMdB, or Stack Exchange/Wikipedia forums) -- but in fact, we will occasionally present an \textit{incongruent} style in the block (e.g., present an impolite Tweet during the polite Tweet block). We expect that incongruency to cause readers to pay more attention to style-specific aspects of the text, as they are unexpected.
We are interested in comparing the eye movements of participants who read a stimulus in the congruent condition with those of participants who read that stimulus in the incongruent condition. Note that the experiment has a between-subjects design, i.e. the same participant does not see the same text in both conditions. 
The congruent reading of the text provides a control.
Figure~\ref{fig:con} shows a concrete example of these two conditions, while Figure~\ref{fig:heatmap} shows a visualization of these contrasted eye movements.

Figure \ref{fig:exp} shows a procedure of our experiments. 
The experimental procedure is as follows (more details in Appendix~\ref{sec:appendix}). Participants complete nine blocks. At the beginning of block, the participant is informed of the style and source, and asked to pay attention to the style of the following texts. Each block contains 10 items, eight of which are \textit{congruent} with the target style. The remaining two items are \textit{incongruent} with the target style. Incongruent items are counterbalanced across participants. Blocks are presented in a random order, and items within the blocks are pseudorandomized to ensure adequate spacing between congruent and incongruent trials~\citep{egner2007congruency} (there is also a \textit{context-free} text as an added control). Participants are asked True/False comprehension questions pseudorandomly after $30\%$ of the items in order to maintain motivation to read carefully. After the experiment concludes, participants complete the Perceived Awareness of Research Hypothesis Scale (PARH) \citep{rubin2016perceived} to evaluate whether demand characterstics \citep{nichols2008good} of the experiment may have influenced reading behavior.
The study procedure was approved by the institutional review board (IRB).

\paragraph{Participants}
We collect data from 20 participants (12 male, 7 female, 1 non-binary; median age 23 years) recruited from the University community and word-of-mouth. An additional 6 participants were recruited but unable to complete the study due to problems with eye calibration. Participants were compensated with a $\$15$ Amazon gift card.

\paragraph{Apparatus}
Monocular eye movement data is collected with an EyeLink 1000 Pro, using the desktop mount, at a rate of 1000Hz. Participants use a chinrest while reading in order to stabilize the head. We use the Experiment Builder software to present stimuli to participants in a 16pt serif font with 1.5 line spacing, on our display monitor with a 508mm display area and a 1680x1050 resolution. Participants are seated with their eyes 50-60cm away from the display monitor.

\paragraph{Study Design Rationale}\label{subsec:rationale} 
Based on the well-documented phenomenon of expectancy effects in cognition (see \citet{schwarz2016rethinking} for further discussion), we assume that the \textit{incongruent} texts that subvert the stylistic expectation will lead to participants reacting with surprise and increased processing difficulty in response to parts of the text associated with the unexpected style. 



Alternative designs that explicitly ask participants to classify an item's style were strongly considered, but were rejected for two reasons: first, we are interested in observing a relatively natural reading process and introducing a classification task runs counter to that goal; second, the style classification task could increase the saliency of not only the target style but also its opposing style, as both can be relevant to the decision (e.g., the presence of an impolite word is relevant to the decision of whether a statement is polite). 
We also considered designs in which congruency is established via explicit text labels rather than implicit expectations, but decided to instead choose an experimental paradigm that adheres as closely as possible to an ecologically valid reading task.

\begin{figure}[t]
    \centering
	\hspace*{-0.3cm}\includegraphics[width=1.1\columnwidth]{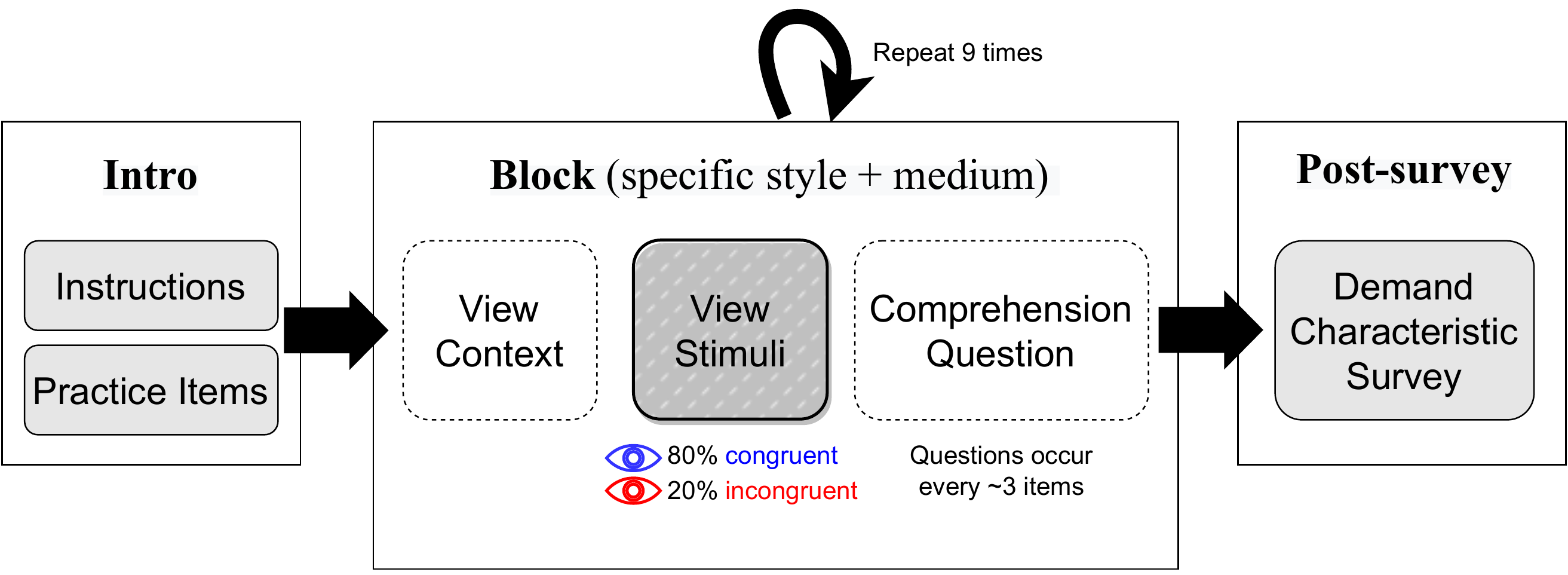}
	\caption{Experimental procedure.
}
	\label{fig:exp}
\end{figure}

\subsection{Pre-processing Eye Tracking Data}
Eye data was delineated into fixations and saccades using the DataViewer software with EyeLink's standard algorithm and default velocity and acceleration thresholds. We further cleaned the data by removing trials with significant track loss (i.e. trials with track loss in over $50\%$ of the text area); $1.5\%$ of trials were removed due to track loss. An outlier analysis showed that $0.5\%$ of fixations were outliers and were removed in our analysis.

\subsection{Calculating Saliency Scores}\label{sec:saliency}
We divide the text into interest areas (IAs) and calculate saliency scores for each IA. We do not segment the IAs such that each IA contains a single word, because in a single fixation people can read a span of about 21 surrounding characters~\cite{rayner1978eye}, meaning that many short words are not fixated on, leading to difficulties with our desired analyses. Instead, we use the natural language processing toolkit (NLTK)'s stopwords list~\citep{bird2009natural} to define each IA such that stopwords share an IA with the closest non-stopword. Specifically, each stopword is combined with the closest non-stopword, with non-stopwords to the right being preferred in the case of a tie. We also ensure that no IA contains a line break.

We utilize two techniques for calculating each eye tracking-based metric for each $\text{IA}_i$. Note that these techniques are applied across all eye tracking measures $x \in \{\text{DT, FRD, GP, DT, RR, PS}\}$ as defined in Section~\ref{subsec:measures}. 
\begin{itemize}[noitemsep,topsep=0pt,leftmargin=7mm]
    \item \textbf{z-score:} For each participant $p_k$, denote the eye tracking measurement in $\text{IA}_i$ as $x_{ki}$. We calculate the participant-specific z-score of eye tracking measurement from $\text{IA}_i$ as $z_k(\text{IA}_i) = \frac{x_{ki} - \mu_k}{\sigma_k}$, where $\mu_k$ and $\sigma_k$ are the participant-specific arithmetic mean and standard deviation, respectively. Then, the saliency score for $\text{IA}_i$ is given by $\frac{\sum_{k = 0}^{n} z_k(\text{IA}_i)}{n}$.
    \item \textbf{raw:} We aggregate the raw values of the eye tracking measurements from each IA. The saliency score for $\text{IA}_i$ is given by $\frac{\sum_{k = 0}^{n}x_{ki}}{n}$.
\end{itemize}

\section{Experimental Results}

\subsection{Comparison with Other Saliency Metrics}

\begin{figure*}
     \centering
     \begin{subfigure}[t]{\columnwidth}
         \centering
         \includegraphics[width=1.2\textwidth]{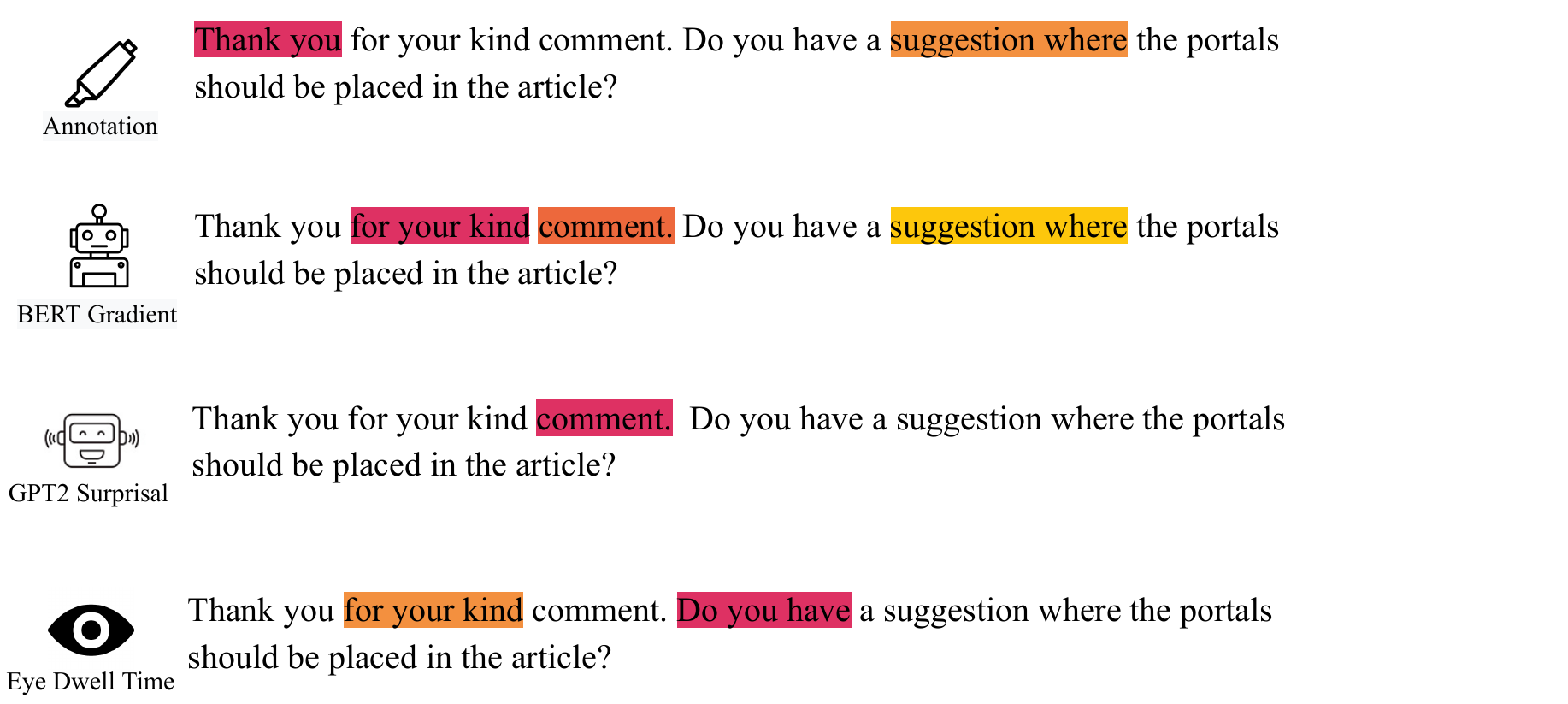}
         \caption{Saliency scores for \textbf{politeness}.}
     \end{subfigure}
      ~ 
     \begin{subfigure}[t]{\columnwidth}
         \centering
         \includegraphics[width=1.2\textwidth]{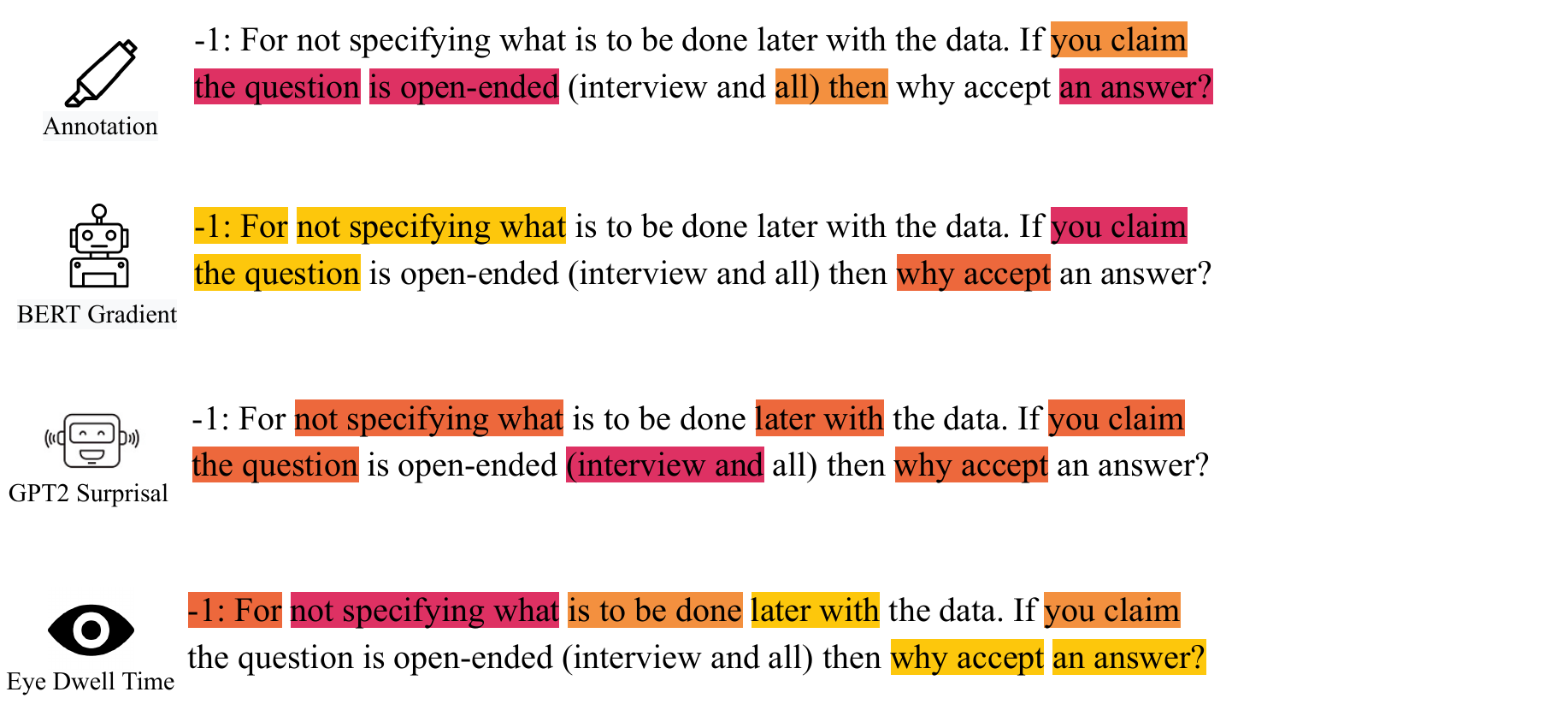}
         \caption{Saliency scores for \textbf{impoliteness}.}
        \label{fig:extra_highlights}
        \end{subfigure}

         \begin{subfigure}[t]{\columnwidth}
         \centering
         \includegraphics[width=\textwidth]{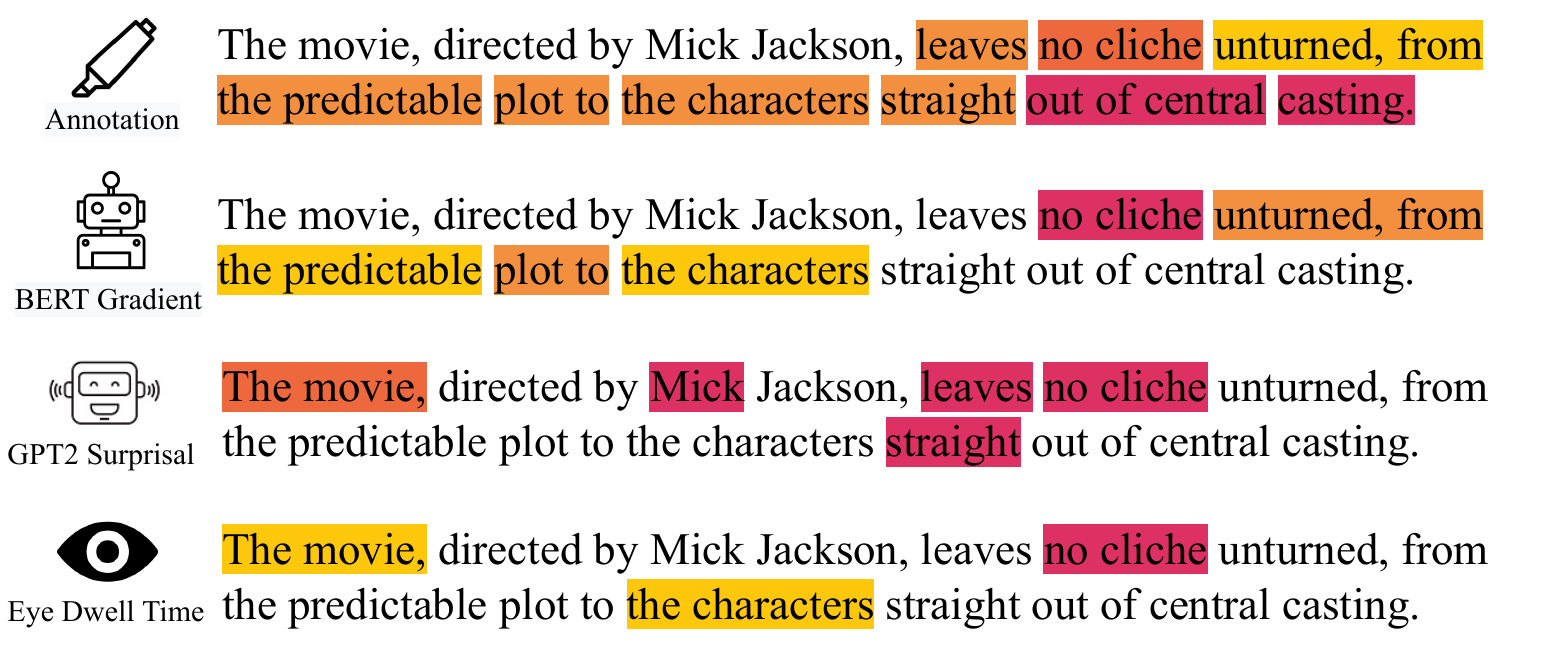}
         \caption{Saliency scores for \textbf{negative} sentiment.}
     \end{subfigure}
     ~
        \begin{subfigure}[t]{\columnwidth}
         \centering
         \includegraphics[width=1.2\textwidth]{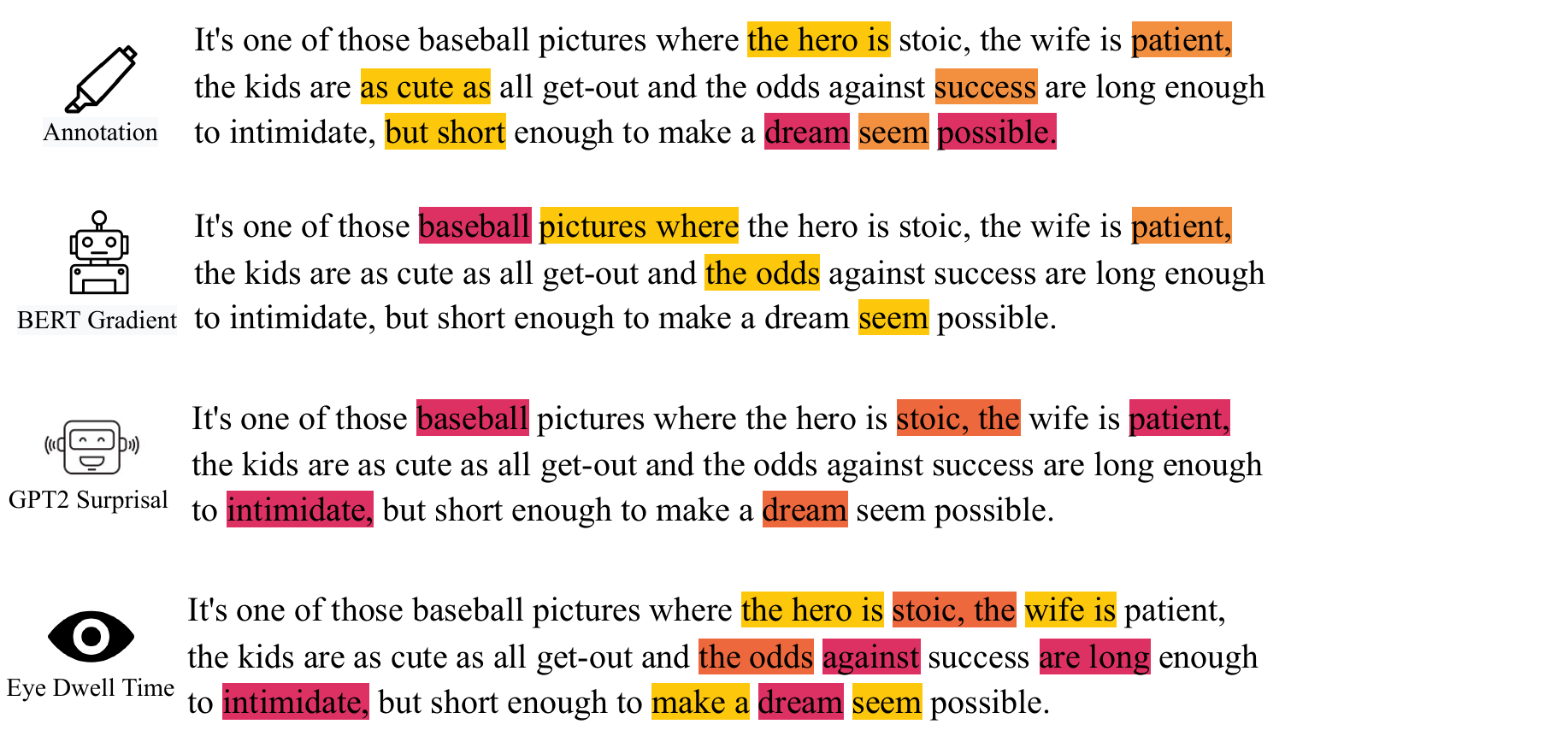}
                  \caption{Saliency scores for \textbf{positive} sentiment.}
        \label{fig:highlighted_words}
               \end{subfigure}
   \caption{A comparison of saliency scores from various methods: manual human annotations, language model introspection, and eye tracking. Darker highlights indicate stronger saliency scores.
   }
	\label{fig:highlighted_words}
\end{figure*}

We investigate how eye tracking metrics compare with other existing measures for lexical-level significance -- namely, human annotations, integrated gradient scores, and large language model surprisal scores  (see Figure~\ref{fig:highlighted_words} for a visualization of these scores):
\begin{itemize}[noitemsep,topsep=0pt,leftmargin=7mm]
    \item \textbf{Surprisal scores}: For the text in the $i$\textsuperscript{th} interest area, denoted $\text{IA}_i$, the surprisal is $P(\text{IA}_i|{\text{IA}_0, \text{IA}_1, ... \text{IA}_{i - 1}})$. We obtain this probability estimate from the pre-trained GPT-2 model \cite{radford2019language}. \footnote{We include word-surprisal scores from GPT-2 as they have previously been found to correlate with human reading times~\cite{wilcox2020predictive}.} In the event that an IA includes multiple tokens, we sum the surprisal of those tokens.
   
    \item \textbf{Model gradient scores}: The integrated gradient method~\cite{sundararajan2017axiomatic} is often used to obtain scores over the input tokens to a deep neural network, where a token's score reflects how much that token influenced the network's final output. We obtain these scores with the Captum codebase~\citep{kokhlikyan2020captum}, using the fine-tuned BERT model from~\citet{hayati2021does}. For $\text{IA}_i$, the integrated gradient score is the average of the individual tokens within $\text{IA}_i$.
    \item \textbf{Human annotations}: Human annotations come from the Hummingbird dataset~\cite{hayati2021does}. Three annotators per item were asked to highlight words that contribute to the text's style.  We averaged these binary highlighting scores over each annotator to arrive at a saliency score for each interest area.
\end{itemize}

\begin{figure*}[t]

    \centering
	\includegraphics[width=0.9\textwidth]{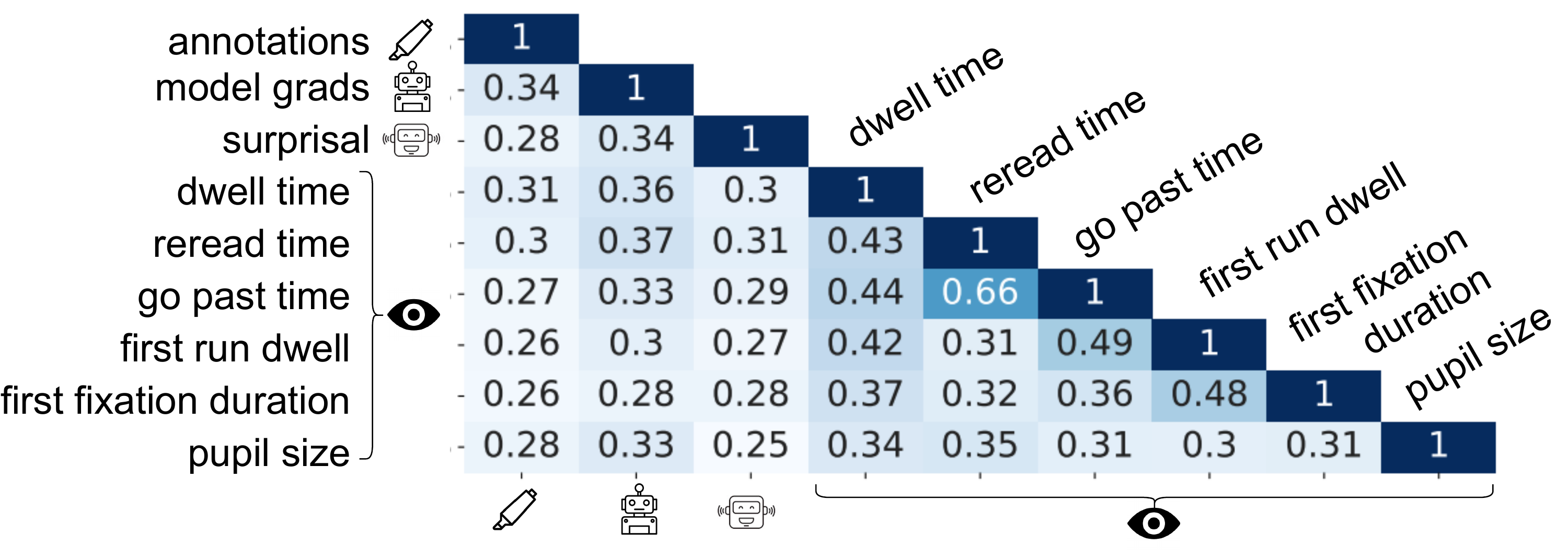}
	\caption{Confusion matrix of the Jaccard similarity score for salient text derived from each metric. (See Appendix for the correlation coefficient for saliency scores derived from each metric.)}
	\label{fig:sim_cor}
\end{figure*}

\begin{figure}[t]
\vspace{-0.75cm}
    \centering
	\includegraphics[width=\columnwidth]{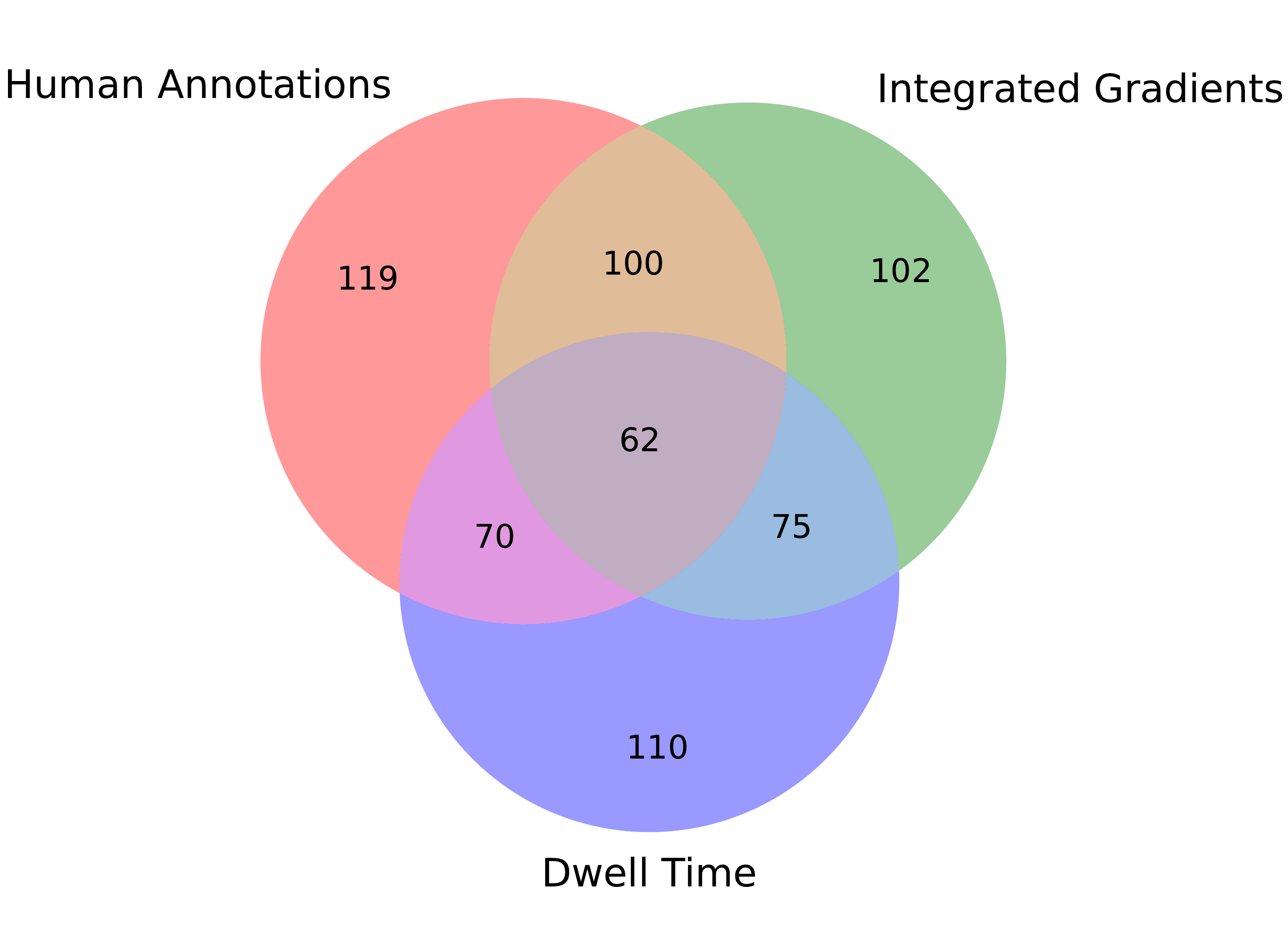}
	\caption{Venn diagram illustrating the intersection of sets of salient interest areas derived from Dwell Time (blue), integrated gradients (green), and human annotations (red).}
	\label{fig:overall_venn}
\end{figure}


Throughout the comparison, we answer the following two questions: How much do the salient IAs derived from each measure overlap and how much does each measure agree on the saliency strength of each IA?

To find the overlap between salient interest areas derived from different measures, we compute a binary saliency map over the dataset for each measure. We then compute the pairwise Jaccard similarity coefficient for each possible pairing of salient text sets (Fig~\ref{fig:sim_cor}), where the Jaccard similarity coefficient is their intersection over union. We use the median saliency score as the threshold that determines whether the IA is labeled ``salient'' so that each measure results in the same number of salient words, allowing a more straightforward comparison between measures.

We find that the intersection over union of salient interest areas from eye tracking methods and both integrated gradient scores and human annotations falls between 0.26 and 0.31. Critically, the three-way intersection over union between salient text from integrated gradients, human annotations, and eye tracking metrics falls between 0.05 and 0.06, indicating that each metric captures a relatively unique set of text within the dataset (see Fig~\ref{fig:overall_venn}).

\begin{figure}[t]
\vspace{-0.2cm}
    \centering
	\includegraphics[width=\columnwidth]{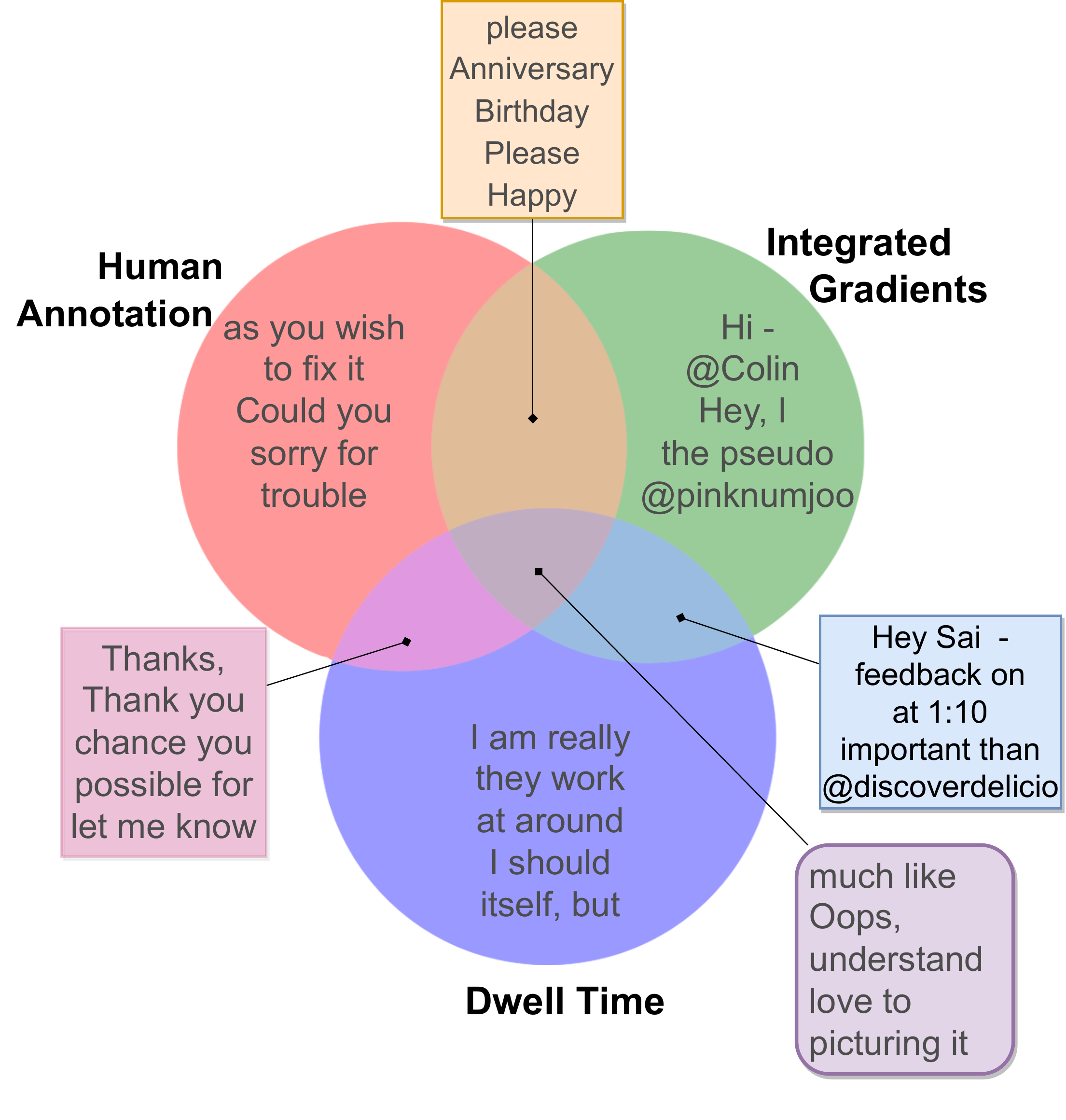}
	\caption{Venn diagram showing interest areas salient to the \textbf{polite} style. For each section of the Venn diagram, the interest areas with the top five highest saliency scores are shown.}
	\label{fig:polite_venn}
\end{figure}

\begin{figure}[t]
    \centering
	\hspace*{-0.2cm}\includegraphics[width=0.51\textwidth,trim=3.5cm 0cm 0cm 0cm,clip]{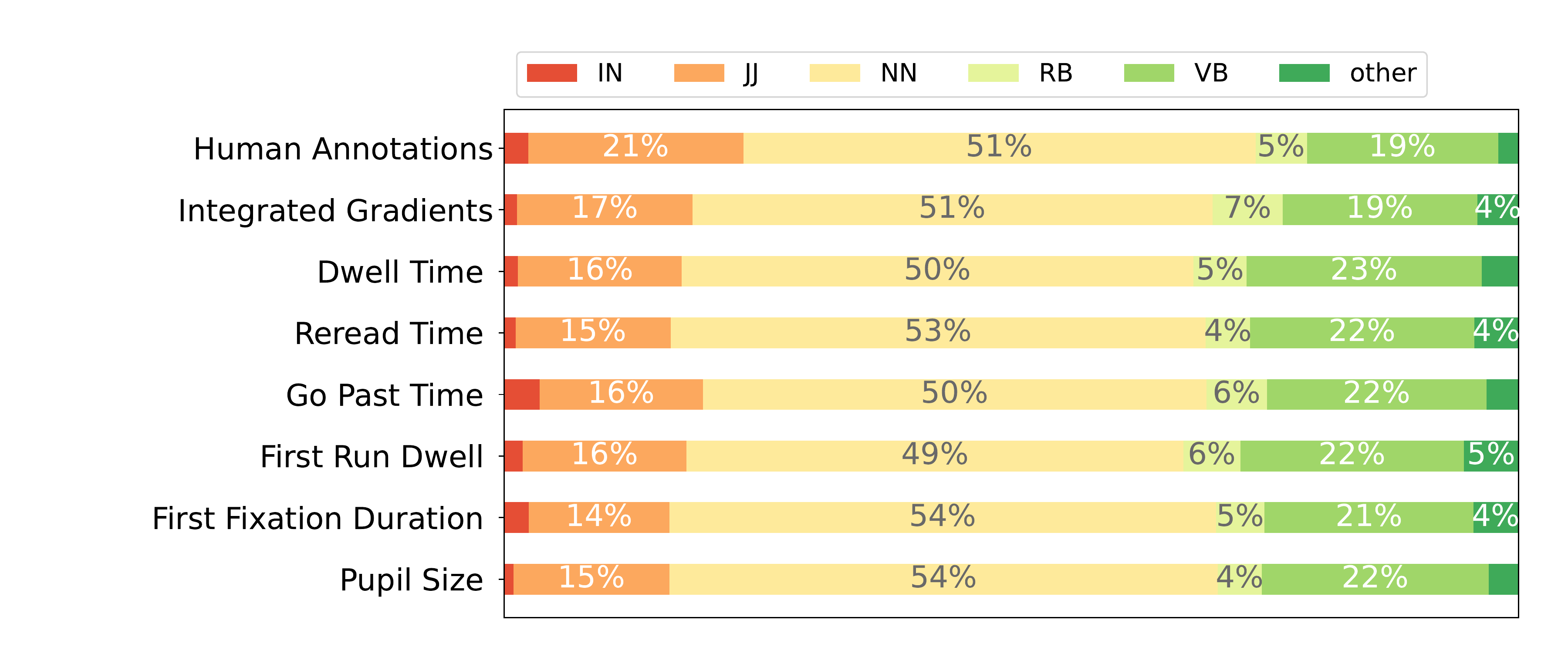}
	\caption{Top 5 most common parts of speech for each measure's salient IA set. IN: prepositions and subordinating conjunctions, JJ: adjectives, NN: nouns, RB: adverbs, VB: verbs.}
	\label{fig:pos}
\end{figure}

We also investigate what types of words are selected as salient by each method by performing part-of-speech (POS) tagging on the salient interest areas for each measure, finding that while distributions of parts of speech are similar, humans select proportionally more adjectives while eye tracking metrics select proportionally more verbs and adverbs (Figure \ref{fig:pos}). This discrepancy may be explained by human annotators focusing more on single words with high stand-alone style (oftentimes these are adjectives such as \textit{happy}, \textit{gracious}), while people's eyes attend to the context surrounding that word (oftentimes this context includes verbs and adverbs). For example, in the polite phrase ``Thank you for removing...,'' human annotators highlight only ``thank you'' whereas eye gaze also focuses on the gerund verb ``removing.''

To measure agreement between different measures with respect to saliency strength, we compute a saliency score for each IA in the dataset derived from each measure. We then compute the pairwise Pearson's \textit{r} correlation coefficient, finding most coefficients are near 0 (see Appendix). In other words, while there is some agreement across human-, machine-, and eye-based methods with respect to which IAs are above median saliency, there is little correlation with respect to the saliency scores themselves.


\subsection{Qualitative Results}
For a qualitative visualization of saliency over the politeness style, see Figure~\ref{fig:polite_venn}. In general, human annotations have a tendency to focus on segments of text with clear style markers. For instance, phrases such as ``please'' are consistently highlighted by human annotators. Our eye tracking data indicates that these phrases do not reliably draw the reader's gaze during the realtime reading process. We notice that the eyes often focus on the object of the politness marker rather than the politeness marker itself: For instance, the polite text ``Thank you for your kind comment,'' human annotators highlight only ``thank you'' whereas gaze data focuses on ``your kind comment.'' 

We also observe that eye data, and in particular dwell time, shows high attention to certain nouns -- i.e., names, usernames, and movie titles. This cannot be explained by word frequency effects, as participants in the control condition did not spend as long attending to these nouns.

\begin{figure}[t]
    \centering
     \includegraphics[width=\columnwidth,trim=2.1cm 0cm 0cm 0cm,clip]{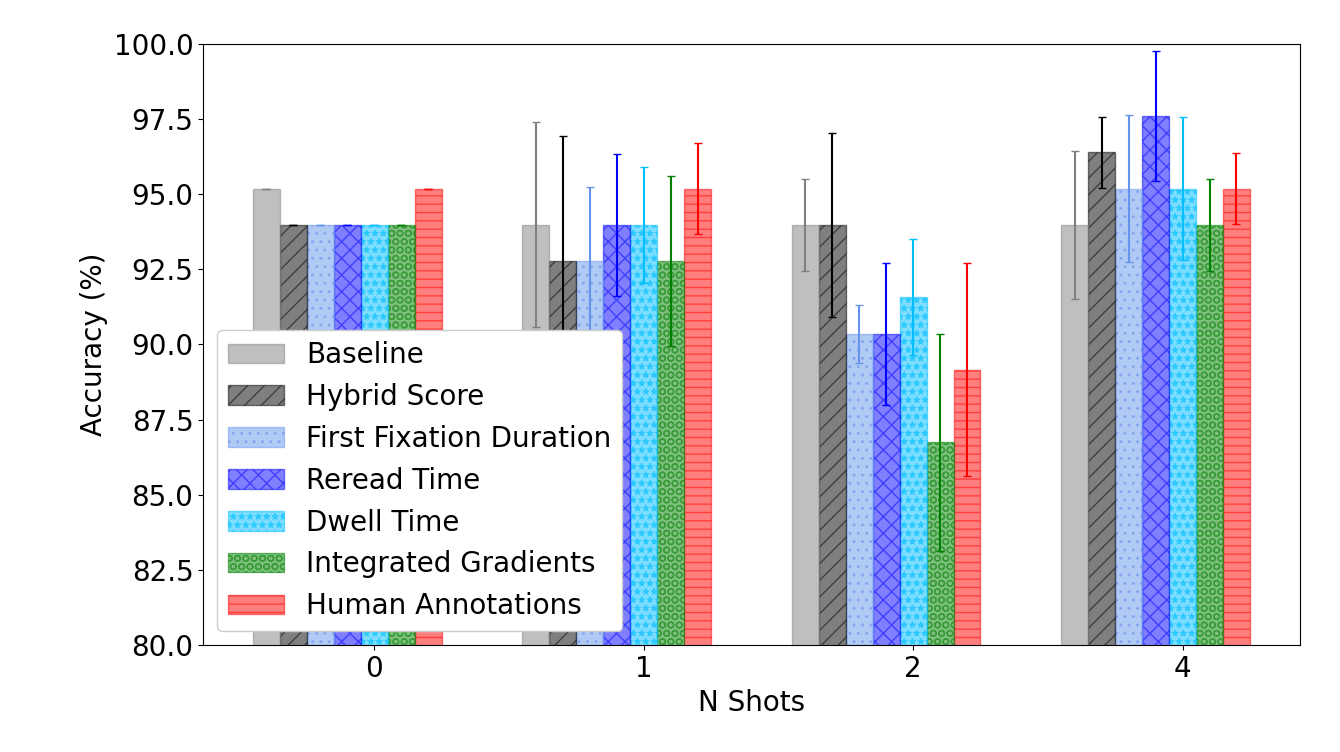}
     \vspace{-2mm}
         \caption{Few-shot learning classification experiment accuracy scores, averaged over 5 rounds with randomly selected demonstrations. Error bars indicate $95\%$ confidence interval.}
	\label{fig:few-shot}
\end{figure}

\subsection{``Eye-in-the-loop'' few-shot learning}
We utilize ``eye-in-the-loop'' few-shot learning in order to roughly probe the cognitive plausibility of GPT-3 \cite{brown2020}. Our prompts present a classification task and include zero to four examples from our dataset, including an ``important words'' section that contains the salient text as defined by each eye-tracking measure, human annotations, and integrated gradient scores (see Section~\ref{sec:saliency} for details). As a baseline, we omit the ``important words.'' We expect that if GPT-3 has a particularly strong cognitive understanding of style processing, ``important words'' from eye movement data may improve its task performance (in these experiments, we use the text-davinci-002 model). Results are relatively inconsistent across each of the four shots, but in most cases, it seems that including salient words has little effect on the model accuracy on the style classification task. A subset of the results are shown in Figure~\ref{fig:few-shot}; see Appendix for full results and prompt details.


\section{Key Findings and Discussion}
Here we discuss the relationship between our results and our research questions:


\textit{ RQ1: Does eye tracking data for saliency meaningfully differ from simply gathering word-level human annotations, or from model-based word importance measures?}
Our data show a substantial difference between eye-tracking-based saliency, model-based saliency, and human annotations. It is perhaps unintuitive that reading behavior would differ from self-reports after reading, but this is consistent with findings in psycholinguistics that establish strong distinctions between explicit measures (i.e., human annotations) and implicit measures (i.e., eye tracking) of human language processing.
Interestingly, there is some intersection between eye tracking-based saliency and model-based saliency that is \textbf{not} shared with human annotators. This suggests that some automatic aspects of human language processing, accessible through eye tracking but not necessarily survey methods, may be shared with large language models.

\textit{RQ2: How can we measure eye movements specific to a high-level textual feature like style, and which eye tracking metrics and data processing methods are best suited to capturing textual saliency?}
The results from our experiment indicate that our experimental paradigm exploiting congruency effects may be effective in finding eye movements specific to certain text features. In a linear mixed effect model analyzing the data, we find significant effects of the congruency condition on dwell time and pupil size (see Appendix \ref{subsec:modeling}). This suggests that the congruency effect does impact reading patterns -- whether this impact is directly linked to the textual style is difficult to definitively answer, but given the overlap between eye-tracking-based style saliency and other style saliency measures, it seems reasonable to believe that the experimental manipulation resulted in an implicit measure of style perception. Experiments based on congruency effects may be a promising route for capturing eye movements related to other high-level textual features such as sarcasm and metaphor. We find that dwell time appears to be the strongest eye-tracking metrics for capturing textual saliency, as it has both the highest overlap with human- and machine-based saliency and most strongly responded to the experimental manipulation. Using the same criteria, we also find that using participant-level z-scores to represent the eye movement data yields the best results.

\section{Limitations}
In this exploratory study, our dataset and sample size are both small, limiting the possibilities for a more thorough evaluation of the data e.g. by fine-tuning a language model. We also note that by design, our experiment presents incongruent items rarely, and consequently we have considerably more congruent datapoints than incongruent datapoints -- an inherent limitation of the proposed experimental paradigm. In light of our results, which suggest that eye-tracking data contains useful and unique information, we plan to develop methods for collecting this kind of real-time human reading data at scale -- i.e., without the constraints of costly in-person eye tracking -- in future work. 

Finally, eye tracking analysis in general is limited by the eye-mind assumption, which holds that the eye fixates on what the mind is currently processing. While there is strong evidence supporting the eye-mind assumption during reading, there is a notable exception: retrieval processes (i.e. accessing memory) are not reflected in eye movements \cite{anderson2004eye}.

\section*{Acknowledgements}
We would like to thank Jeffrey Bye, Andrew Elfenbein, Charles Fletcher, Shirley Hayati, Brooke Lea, Andreas Schramm, and Mariya Toneva for their valuable feedback and insightful suggestions regarding the experimental procedure and data analysis.
We would also like to thank Miguel Miguelez Diaz, Risako Owan, Faziel Khan, and Josh Spitzer-Resnick for testing and critiquing the initial experimental pipeline.
This research received funding from the Sony Research Innovation Award.

\bibliography{text-style.bib,anthology,custom}
\bibliographystyle{acl_natbib}

\appendix
\clearpage
\section{Appendix}
\label{sec:appendix}
\subsection{Experimental Materials}
\begin{figure*}
     \centering
          \begin{subfigure}[t]{0.4\textwidth}
         \centering
         \frame{\includegraphics[width=\textwidth]{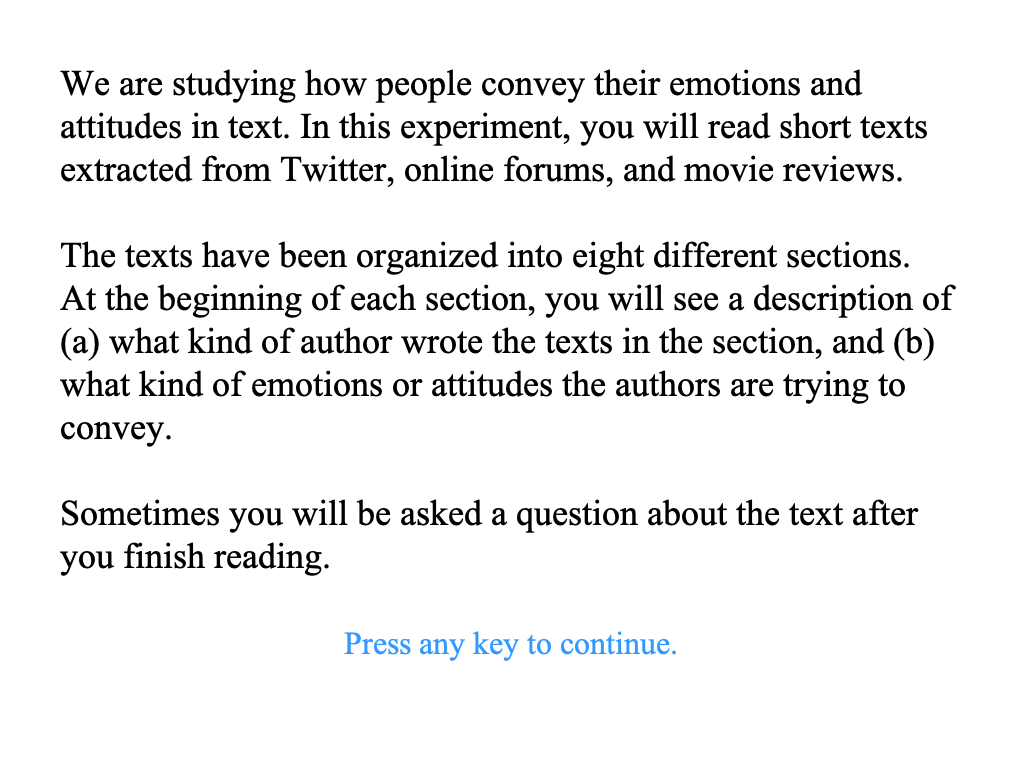}}
         \caption{Experiment instructions screen.}
     \end{subfigure}
     \hfill
     \begin{subfigure}[t]{0.4\textwidth}
         \centering
         \frame{\includegraphics[width=\textwidth]{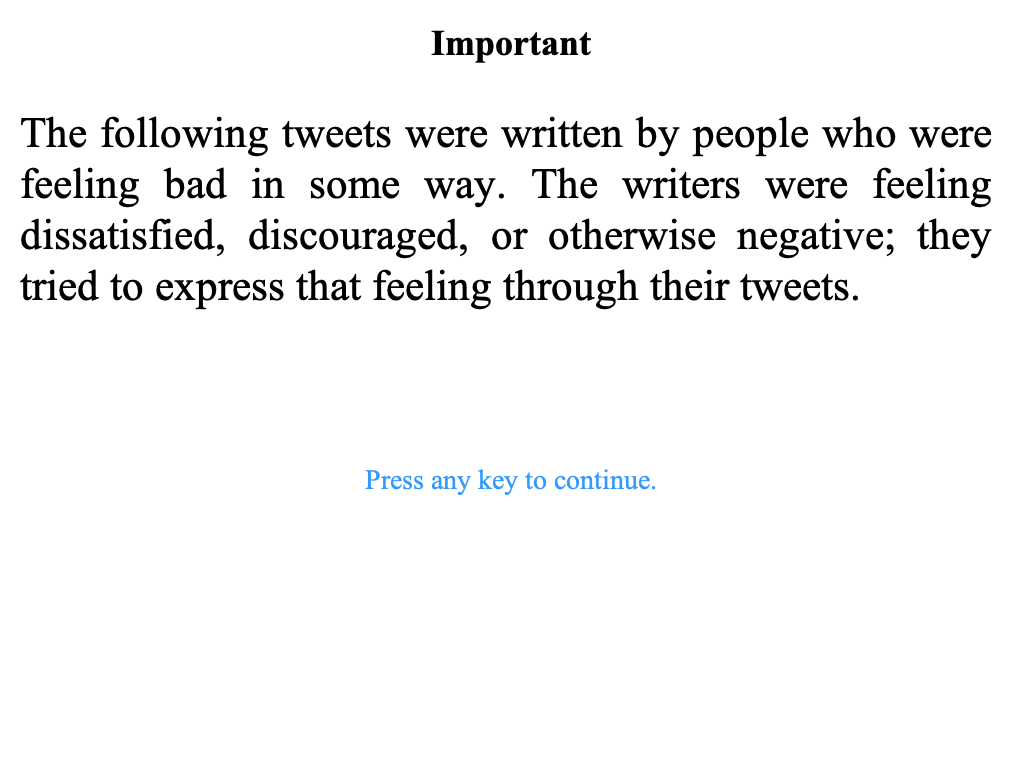}}
         \caption{One of the ``context'' screens shown at the beginning of each block. This information makes participants aware of what type of text to expect in the following screens.}
     \end{subfigure}
     \hfill
     \begin{subfigure}[t]{0.4\textwidth}
         \centering
         \frame{\includegraphics[width=\textwidth]{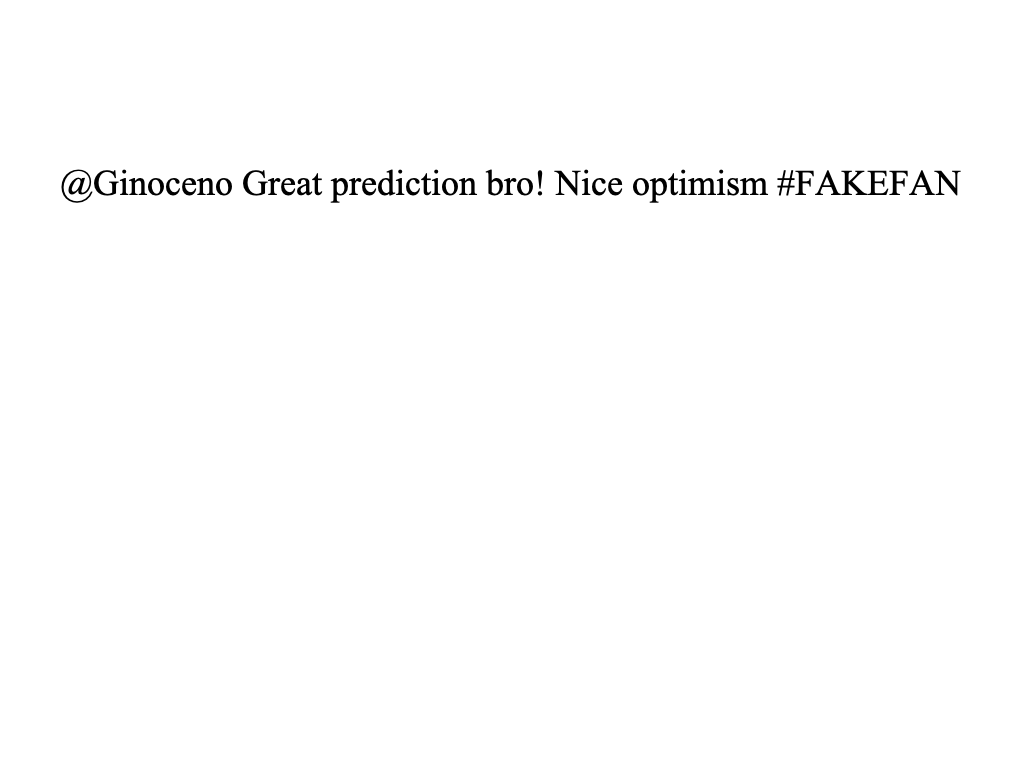}}
         \caption{One of the screens displaying an item from the dataset.}
     \end{subfigure}
     \hfill
     \begin{subfigure}[t]{0.4\textwidth}
         \centering
         \frame{\includegraphics[width=\textwidth]{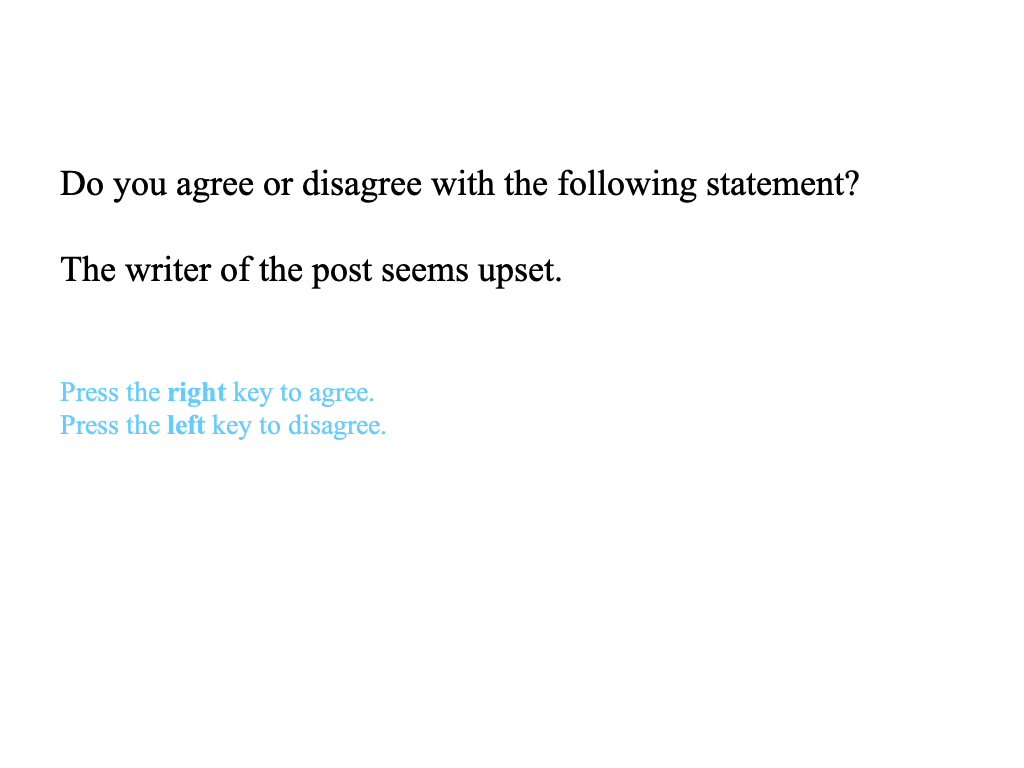}}
         \caption{One of the comprehension question screens.}
     \end{subfigure}
        \caption{Screenshots from the experiment program.}
        \label{fig:experiment_screenshots}
\end{figure*}

The following materials were presented to participants during the experiment. Informed consent was obtained from each participant before the experiment began. Instructions were displayed as shown in Figure~\ref{fig:experiment_screenshots}. 

The practice items, which participants completed after reading the instructions and before beginning the experiment, were as follows:
\begin{lstlisting}
    Text: What does this have to do with programming?  Are you trying to solve this problem with a program?
    Question: None

    Text: this is source code... what is the question? Do you really think that throwing code at us will solve your problem?!
    Question: Do you agree or disagree with the following statement: The writer of the post seems upset.
\end{lstlisting}

See also Figure ~\ref{fig:experiment_screenshots} for screenshots of the display shown to participants at various points in the experiment.

\subsection{Mixed Effect Modeling}\label{subsec:modeling}
We fit linear mixed effect models to predict our eye tracking measures, using the R packages lme4 and lmetest. Our fixed effects are the number of characters in the interest area, the HAL frequency of the interest area, whether the previous interest area was viewed, and whether the interest area is in the congruent or incongruent condition. Our random effect is the participant ID. All variables are normalized prior to analysis.
\begin{lstlisting}
    model = lmer(EYE_TRACKING_MEASURE ~ 1 + congruent + previous_viewed+ LENGTH + HAL_FREQ + (1 | RECORDING_SESSION_LABEL))
\end{lstlisting}

The Dwell Time and Pupil Size eye tracking measure showed significance for the the fixed congruency effect. The other eye tracking measures -- First Run Dwell Time, First Fixation Duration, Reread Time, and Go Past Time -- result in a singular fit, likely because they are considerably more sparse (i.e., many interest areas have a null values for these metrics).

\begin{table}[h]
  \centering
  \begin{tabular}{lrrrr}
  \hline
  & t value & $Pr(>|t|)$ & Sig. & VIF \\
  \hline
  (Intercept) & -19.114 & < 0.001 & $\ast\ast\ast$ &  \\
  frequency & -18.238 & < 0.001 & $\ast\ast\ast$ & 2.53\\
  length & 31.858 & < 0.001 & $\ast\ast\ast$ & 2.53 \\
  congruency & 2.449 & < 0.05 & $\ast$ & 1.00\\
  previous IA & 26.662 & <0.001 & $\ast$ & 1.00\\
  \hline
  \end{tabular}
  \caption{Fixed Effects: predicting dwell time}
  \label{tab:third}
  \end{table}

\begin{table}[h]
  \centering
  \begin{tabular}{lrrrr}
  \hline
  & t value & $Pr(>|t|)$ & Sig. & VIF \\
  \hline
  (Intercept) & -4.098 & < 0.001 & $\ast\ast\ast$ &  \\
  frequency & 1.865 & 0.06 & . & 2.28 \\
  length & 3.056 & < 0.01 & $\ast\ast$ & 2.27 \\
  congruency & -8.382 & < 0.001 & $\ast\ast\ast$ & 1.00\\
  previous IA & 9.915 & <0.001 & $\ast\ast\ast$ & 1.00\\
  \hline
  \end{tabular}
  \caption{Fixed Effects: predicting pupil size}
  \label{tab:third}
  \end{table}
  
We tested variables for collinearity using the variance inflation factor (VIF) \citep{zuur2010protocol} (none exceeded the recommended threshold of 3).

\subsection{Additional Saliency Comparisons}
\subsubsection{Saliency Scores}
\begin{figure*}
    \centering
    \includegraphics[width=2\columnwidth]{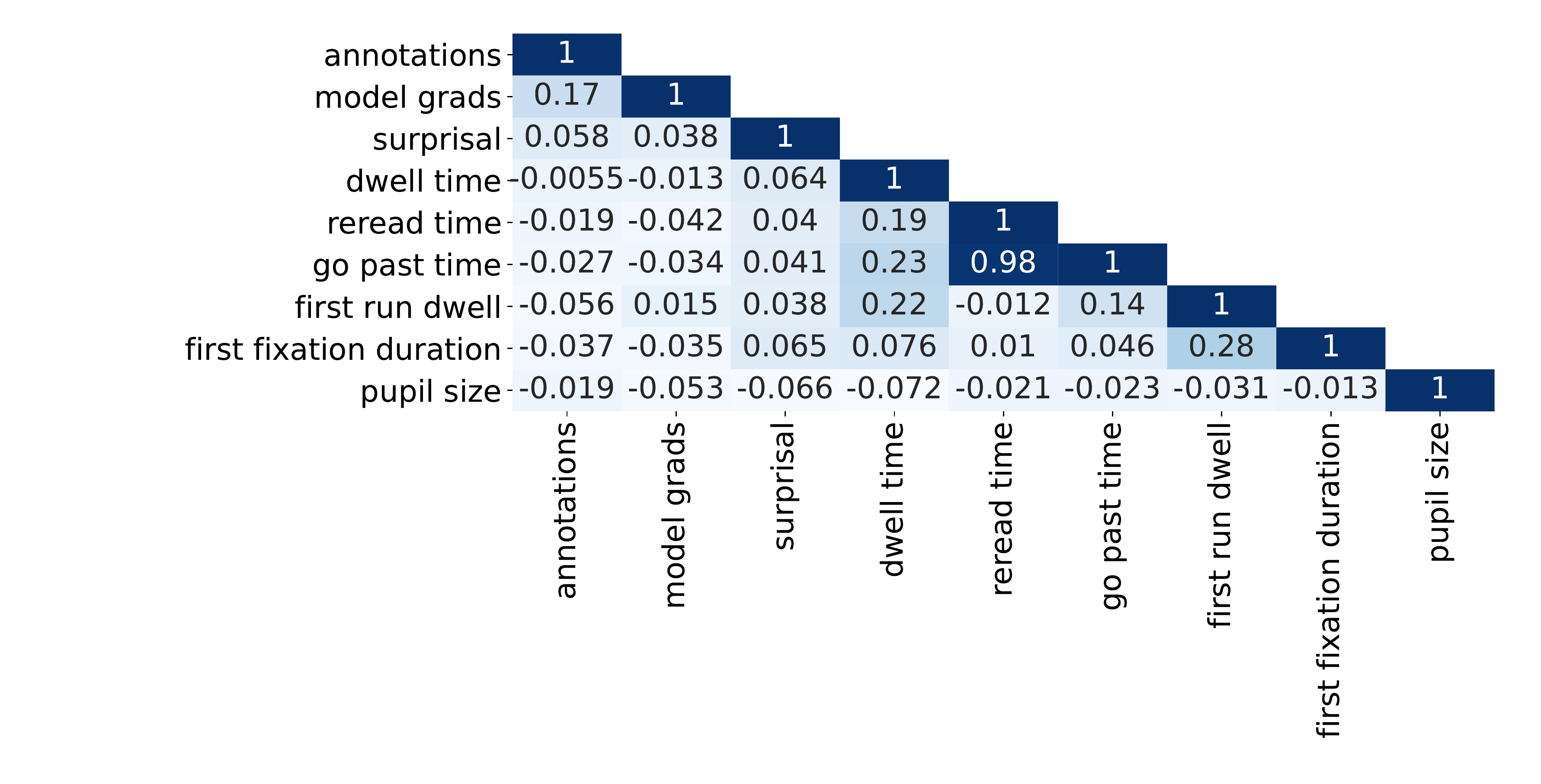}
    \caption{Correlations (Pearson's r) between the saliency scores derived from each method.}
    \label{fig:correlations}
\end{figure*}

Figure ~\ref{fig:correlations} shows the Pearson's $r$ value for saliency score over interest areas derived from each method. We also include more example items from the dataset with associated saliency scores in Figure~\ref{fig:extra_highlights}.

\subsection{Few-Shot Learning Experiment Details and Results}
The full few-shot learning results can be found in Table~\ref{tab:fewshotfull}. The experiment was conducted with the OpenAI API\footnote{https://openai.com/api, accessed in accordance with OpenAI's terms of use} completion endpoint and the following parameters: the text-davinci-002 model, a temperature of 0, and a top\_p of 1.

We generated in-context learning prompts over our dataset by including \textit{important words} as follows:
\begin{lstlisting}
Decide whether the following text is Polite or Impolite.
Text: Thank you for your kind comment. Do you have a suggestion where the portals should be placed?
Important words: thank you, suggestion
Polite or Impolite:
\end{lstlisting}

\begin{table*}[]
\resizebox{\textwidth}{!}{
\begin{tabular}{l|l|l|l|l|l|l}
Metric for Saliency & Data aggregation (eye-tracking only) & Experimental Conditions & 0-shot & 1-shot  & 2-shot & 4-shot \\\hline
Baseline & NA & NA & 95.18 & 93.98 (2.46) & 90.36 (0.96) & 95.18 (0.96) \\
Human Annotations & NA & NA & 93.98 & 91.57 (2.89) & 90.36 (3.27) & 93.98 (1.80) \\
Integrated Gradients & NA & NA & 93.98 & 93.98 (1.93) & 92.77 (2.46) & 96.39 (0.96) \\
GPT2 Surprisal & NA & NA & 93.98 & 92.77 (0.96) & 92.77 (0.96) & 97.59 (1.18) \\
Dwell Time & z score & All & 93.98 & 92.77 (1.80) & 93.98 (0.96) & 96.39 (2.89) \\
Dwell Time & z score & Incongruent - Congruent & 93.98 & 93.98 (1.93) & 91.57 (1.93) & 95.18 (2.36) \\
Dwell Time & LME & All & 93.98 & 92.77 (0.96) & 92.77 (0.96) & 97.59 (1.18) \\
Dwell Time & LME & Incongruent - Congruent & 93.98 & 92.77 (1.18) & 91.57 (1.93) & 95.18 (2.36) \\
Dwell Time & raw & All & 93.98 & 93.98 (1.18) & 95.18 (1.93) & 96.39 (1.80) \\
Dwell Time & raw & Incongruent - Congruent & 93.98 & 92.77 (1.80) & 89.16 (2.89) & 95.18 (2.46) \\
Reread Time & z score & All & 93.98 & 92.77 (0.96) & 92.77 (0.96) & 97.59 (1.18) \\
Reread Time & z score & Incongruent - Congruent & 93.98 & 93.98 (2.36) & 90.36 (2.36) & 97.59 (2.16) \\
Reread Time & raw & All & 93.98 & 92.77 (1.18) & 91.57 (2.46) & 93.98 (2.81) \\
Reread Time & raw & Incongruent - Congruent & 92.77 & 92.77 (2.46) & 86.75 (2.81) & 96.39 (1.80) \\
Go Past Time & z score & All & 93.98 & 92.77 (0.96) & 92.77 (0.96) & 97.59 (1.18) \\
Go Past Time & z score & Incongruent - Congruent & 93.98 & 91.57 (2.89) & 87.95 (3.86) & 92.77 (2.46) \\
Go Past Time & raw & All & 92.77 & 92.77 (0.96) & 91.57 (2.46) & 96.39 (1.18) \\
Go Past Time & raw & Incongruent - Congruent & 93.98 & 92.77 (3.27) & 90.36 (3.20) & 93.98 (2.46) \\
First Run Dwell Time & z score & All & 93.98 & 92.77 (0.96) & 92.77 (0.96) & 97.59 (1.18) \\
First Run Dwell Time & z score & Incongruent - Congruent & 93.98 & 92.77 (2.46) & 92.77 (1.18) & 92.77 (2.46) \\
First Run Dwell Time & raw & All & 93.98 & 92.77 (1.18) & 92.77 (2.46) & 96.39 (2.36) \\
First Run Dwell Time & raw & Incongruent - Congruent & 93.98 & 92.77 (1.80) & 89.16 (3.54) & 93.98 (2.46) \\
First Run Dwell Time & LME & All & 93.98 & 92.77 (1.93) & 92.77 (2.36) & 95.18 (2.46) \\
First Run Dwell Time & LME & Incongruent - Congruent & 93.98 & 92.77 (2.46) & 89.16 (3.54) & 93.98 (2.46) \\
First Fixation Duration & z score & All & 93.98 & 92.77 (0.96) & 92.77 (0.96) & 97.59 (1.18) \\
First Fixation Duration & z score & Incongruent - Congruent & 93.98 & 92.77 (2.46) & 90.36 (0.96) & 95.18 (2.46) \\
First Fixation Duration & raw & All & 93.98 & 93.98 (2.64) & 89.16 (1.80) & 96.39 (0.96) \\
First Fixation Duration & raw & Incongruent - Congruent & 93.98 & 92.77 (2.81) & 90.36 (3.27) & 95.18 (1.93) \\
Pupil Size & z score & All & 93.98 & 92.77 (0.96) & 92.77 (0.96) & 97.59 (1.18) \\
Pupil Size & z score & Incongruent - Congruent & 92.77 & 93.98 (2.46) & 86.75 (4.31) & 93.98 (1.80) \\
Pupil Size & raw & All & 93.98 & 92.77 (1.18) & 92.77 (1.80) & 95.18 (2.81) \\
Pupil Size & raw & Incongruent - Congruent & 93.98 & 91.57 (1.93) & 86.75 (4.03) & 96.39 (2.46) \\
Pupil Size & LME & All & 93.98 & 91.57 (2.46) & 95.18 (2.36) & 92.77 (1.18) \\
Pupil Size & LME & Incongruent - Congruent & 93.98 & 92.77 (2.16) & 86.75 (4.03) & 96.39 (2.46) \\
Hybrid (Human + Dwell Time) & z score & All & 95.18 & 93.98 (1.18) & 93.98 (2.46) & 96.39 (1.52) \\
Hybrid (Human + Dwell Time) & z score & Incongruent - Congruent & 93.98 & 92.77 (4.15) & 93.98 (3.05) & 96.39 (1.18)
\end{tabular}
}
\caption{Accuracy results on few-shot learning experiments over dataset. For 1-, 2-, and 4-shot learning, five different randomly selected prompts were chosen and the average accuracy is reported (the $95\%$ confidence interval is reported in parentheses after the accuracy score).}
\label{tab:fewshotfull}
\end{table*}

\end{document}